\documentclass[letterpaper, 10pt, conference]{ieeeconf}

\IEEEoverridecommandlockouts
\usepackage{amsmath,amssymb,amsfonts}
\usepackage{booktabs}
\usepackage{algorithm}
\usepackage{algpseudocode}
\usepackage{graphicx}
\usepackage{bm}
\usepackage{cite}
\usepackage{subcaption}
\usepackage{multirow}
\usepackage{url}
\usepackage{xcolor}
\usepackage{array}
\let\labelindent\relax
\usepackage{enumitem}
\usepackage{cuted}

\usepackage{etoolbox}

\newif\ifEGTtightspacing
\EGTtightspacingtrue

\newcommand*{\EGTdisplaySpacing}{%
  \ifEGTtightspacing
    \setlength{\abovedisplayskip}{3pt plus 1pt minus 0.5pt}%
    \setlength{\belowdisplayskip}{3pt plus 1pt minus 0.5pt}%
    \setlength{\belowdisplayshortskip}{2pt plus 1pt minus 0.5pt}%
    \setlength{\jot}{1.5pt}%
  \else
    \setlength{\abovedisplayskip}{4.5pt plus 1pt minus 1pt}%
    \setlength{\belowdisplayskip}{4.5pt plus 1pt minus 1pt}%
    \setlength{\belowdisplayshortskip}{3pt plus 1pt minus 1pt}%
    \setlength{\jot}{2pt}%
  \fi
  \setlength{\abovedisplayshortskip}{0pt plus 1pt}%
}

\AtBeginDocument{%
  \ifEGTtightspacing
    \setlength{\textfloatsep}{7pt plus 1pt minus 1pt}%
    \setlength{\dbltextfloatsep}{7pt plus 1pt minus 1pt}%
    \setlength{\intextsep}{7pt plus 1pt minus 1pt}%
    \setlength{\floatsep}{5pt plus 1pt minus 1pt}%
    \setlength{\dblfloatsep}{5pt plus 1pt minus 1pt}%
    \setlength{\stripsep}{4pt plus 1pt minus 1pt}%
  \else
    \setlength{\textfloatsep}{9pt plus 2pt minus 1pt}%
    \setlength{\dbltextfloatsep}{9pt plus 2pt minus 1pt}%
    \setlength{\intextsep}{8pt plus 2pt minus 1pt}%
    \setlength{\floatsep}{6pt plus 1pt minus 1pt}%
    \setlength{\dblfloatsep}{6pt plus 1pt minus 1pt}%
    \setlength{\stripsep}{5pt plus 1pt minus 1pt}%
  \fi
  \setlist[itemize]{topsep=2pt,itemsep=0.5pt,parsep=0pt,partopsep=0pt}%
  \setlist[enumerate]{topsep=2pt,itemsep=1pt,parsep=0pt,partopsep=0pt}%
  \captionsetup[figure]{skip=3pt}%
  \captionsetup[table]{skip=3pt,position=top}%
  \captionsetup[subfigure]{skip=2pt}%
}

\makeatletter
\let\EGT@originalstartsection\@startsection
\def\@startsection#1#2#3#4#5#6{%
  \ifnum#2<3\relax
    \ifEGTtightspacing
      \def\EGT@nextsection{\EGT@originalstartsection{#1}{#2}{#3}%
        {\glueexpr#4*2/3\relax}{\glueexpr#5*3/4\relax}{#6}}%
    \else
      \def\EGT@nextsection{\EGT@originalstartsection{#1}{#2}{#3}%
        {\glueexpr#4*3/4\relax}{\glueexpr#5*4/5\relax}{#6}}%
    \fi
  \else
    \def\EGT@nextsection{\EGT@originalstartsection{#1}{#2}{#3}{#4}{#5}{#6}}%
  \fi
  \EGT@nextsection
}
\makeatother

\newcommand*{\EGTsmallDisplay}{%
  \edef\EGTrestoreDisplaySpacing{%
    \baselineskip=\the\baselineskip\relax
    \lineskip=\the\lineskip\relax
    \lineskiplimit=\the\lineskiplimit\relax
    \abovedisplayskip=\the\abovedisplayskip\relax
    \belowdisplayskip=\the\belowdisplayskip\relax
    \abovedisplayshortskip=\the\abovedisplayshortskip\relax
    \belowdisplayshortskip=\the\belowdisplayshortskip\relax
  }%
  \small
  \EGTrestoreDisplaySpacing
  \EGTdisplaySpacing
}
\AtBeginEnvironment{equation}{\EGTsmallDisplay}
\AtBeginEnvironment{equation*}{\EGTsmallDisplay}
\AtBeginEnvironment{align}{\EGTsmallDisplay}
\AtBeginEnvironment{align*}{\EGTsmallDisplay}

\title{\LARGE\bfseries
EGT: Execution-Grounded Traversability through Learned Closed-Loop Trajectory Distributions
}

\author{Chinmay Mundane$^{1}$, Amith Manoharan$^{1}$, and Arun Kumar Singh$^{1}$
\thanks{$^1$ are with the Institute of Technology, University of Tartu, Estonia.}
}

\begin{document}

\maketitle
\thispagestyle{empty}
\pagestyle{empty}


\begin{strip}
    \centering
    \begin{minipage}{\textwidth}
        \centering
        \includegraphics[width=\textwidth]{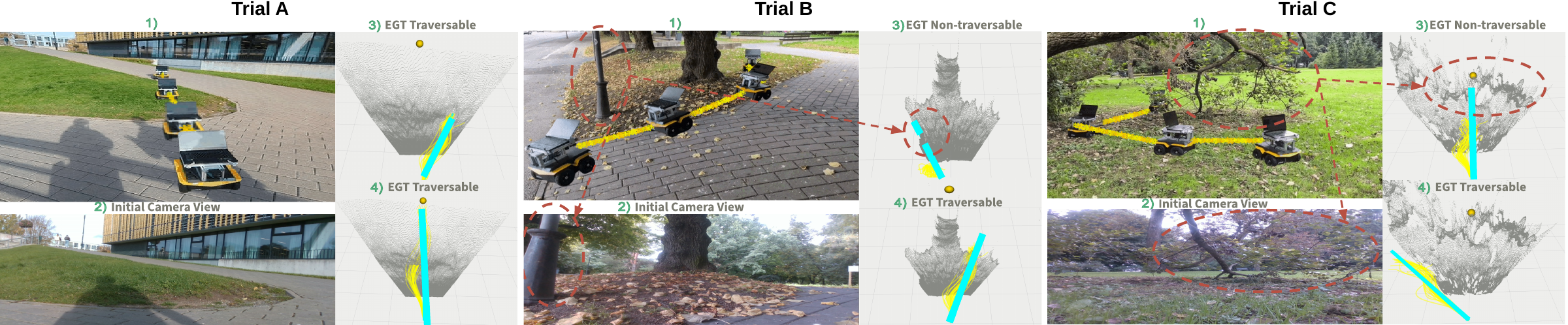}
        \captionof{figure}{\footnotesize{Real-world EGT tests across three terrain trials. Each trial shows (1) the robot motion, (2) the initial camera view, and (3)–(4) the candidate reference paths (cyan) with their FM-predicted closed-loop trajectories (yellow). The accompanying video shows the subsequent replanning steps. Trial A: Due to the small slope angle, the predicted trajectories closely follow both cyan paths to their endpoints, so EGT classifies both paths as traversable, further confirmed by the actual vehicle motion. Trial B: One candidate path hits a lamp post, and the predicted trajectory stops early. EGT classifies this blocked path as non-traversable (3) and the path that moves away from the tree as traversable (4) Trial C: A straight candidate path hits a low tree branch. EGT classifies this blocked path as non-traversable (3) and the path that steers away from the branch as traversable (4). }}
        \label{fig:teaser_overview}
    \end{minipage}
\end{strip}

\begin{abstract}

Rough terrain traversability is often estimated through point-wise roll, pitch, or force-angle margins along a planned path. However, these metrics are often inaccurate and overly conservative because they ignore how terrain disturbances and low-level controller corrections can affect the vehicle's motions. To overcome these limitations, we introduce \emph{Execution-Grounded Traversability} (EGT), defined as the probability that a vehicle--controller pair can execute a candidate path while meeting trajectory-level tracking and progress requirements.
We estimate EGT by predicting the distribution of closed-loop trajectories using a latent-space flow matching (FM) model conditioned on multi-resolution terrain features and 3D vehicle attitude and wheel-ground contact points prior. Across 650 simulated trials on unseen terrains, EGT reduces safe-path rejection to 2.2\% (versus 11.4\%--39.5\% for pose-margin and force-angle baselines) and unsafe acceptance to 7.7\%.
Integrated into a receding-horizon sampling-based planner, EGT reduces navigation failure rates by 24--42 and 30--44 percentage points over~\cite{leininger2024gaussian} and~\cite{tal}, respectively, while cutting goal-reaching time by up to 20.5\% and 45\%.
Finally, we demonstrate zero-shot transfer of our simulation-trained model to real-world outdoor experiments.

\end{abstract}

\section{Introduction}
\label{intro}
Autonomous wheeled vehicles used outdoors for applications such as search and rescue, construction, and forestry must accurately evaluate whether they can follow a reference path over rough terrain without tipping over or getting stuck. Standard traversability methods evaluate safety along this path using tip-over stability margins~\cite{Papadopoulos01012000} or limits on terrain-induced roll and pitch angles~\cite{takemura2022computationally,krusi2017driving}. However, relying on attitude limits or tip-over margins to evaluate traversability---even dynamic metrics like the force-angle measure~\cite{Papadopoulos01012000}---tends to be both overly conservative and inaccurate for three key reasons:
\begin{enumerate}[leftmargin=*]
    \item \textbf{Pointwise limits ignore closed-loop recovery (overly conservative):} Checking stability or attitude limits at every point along a path penalizes brief posture excursions. A large transient roll or pitch angle does not make a path infeasible if the vehicle's momentum and low-level controller can ensure a safe recovery.
    \item \textbf{Rotational stability does not guarantee traversal (inaccurate):} Force-angle and attitude limits only check for rotational tip-over.
    Even when a vehicle stays within safe posture limits and tip-over margins, wheel slip can stop forward progress or push the vehicle sideways into obstacles or ditches.
    \item \textbf{Reference-path checks ignore closed-loop deviation (inaccurate):} Calculating posture or tip-over margins along the reference path may provide limited insight because terrain disturbances and subsequent controller corrections can make the vehicle depart substantially from the planned reference.
\end{enumerate}

\emph{Therefore, in this paper, we do not define traversability from terrain and posture features alone; instead, we evaluate whether a specific vehicle--controller pair can execute a reference path under closed-loop control.}
Formally, we define \emph{Execution-Grounded Traversability} (EGT) as the probability that a vehicle--controller pair can execute a candidate reference path while keeping the path-tracking error below a set limit and completing a required fraction of its arc length.
This execution-based perspective reshapes how we factor vehicle stability into traversability assessments. Specifically, our core insight is that physical failures---like tipping into a ditch or stalling on a steep slope---directly derail path tracking or halt forward progress. Conversely, satisfying attitude or force-angle bounds alone does not guarantee that the vehicle will complete the path.

To compute EGT, we must predict the distribution of closed-loop path-tracking behavior on rough terrain. One approach, proposed by Wang et al.~\cite{pmlr-v164-wang22c}, learns an open-loop, action-conditioned dynamics model and pairs it with a tracking controller at inference time. In effect, the learned model acts as a simulator, generating closed-loop executions through repeated evaluations of the dynamics model and a low-level controller. However, this recursive rollout compounds prediction errors: an inaccurate state prediction alters the controller's corrective action, which further degrades subsequent predictions. Moreover, these corrections can push the model toward state--action pairs that are poorly represented in the training data, further reducing accuracy. Finally, and most importantly, their learned target is action-conditioned vehicle dynamics,
rather than the complete execution distribution conditioned on a reference path.

In this paper, we bypass the recursive rollouts of~\cite{pmlr-v164-wang22c} by treating the vehicle and its feedback controller as a single closed-loop system. Rather than simulating step-by-step dynamics, we directly learn the distribution of realized trajectories from closed-loop executions in simulation. We model this conditional distribution using a latent-space flow matching (FM) architecture conditioned on multi-resolution terrain features and a 3D kinematic pose--contact prior. Given this context and a candidate path, our learned
FM model jointly predicts the entire sequence of
closed-loop waypoints, without recursive evaluations
of vehicle dynamics and a tracking controller.

Our EGT method provides two main advantages.
First, across 650 simulated trials on unseen terrains, EGT reaches 97.4\% classification accuracy.
It is much less conservative and more accurate than the force-angle measure~\cite{Papadopoulos01012000} and roll/pitch limit methods~\cite{takemura2022computationally,krusi2017driving}, cutting the conservative rejection of safe paths from 11.4\%--39.5\% down to 2.2\% and unsafe acceptance down to 7.7\%. Second, when we use EGT with conditional value-at-risk (CVaR) execution costs in a simple receding-horizon sampling-based planner, we reduce failure rates by 24--42 percentage points over~\cite{leininger2024gaussian} and 30--44 percentage points over~\cite{tal}.
At the same time, our planner reduces mean goal-reaching time by up to 20.5\% and 45\% over these baselines.

\noindent In summary, the contributions of our work are:

\begin{itemize}[leftmargin=*]
    \item We introduce EGT, a probabilistic traversability
    metric based on closed-loop execution requirements.

    \item We develop a closed-loop trajectory predictor based on latent FM that allows us to estimate EGT along a reference path.

    \item We combine EGT with risk-aware planning and
    demonstrate improvements in path classification
    and navigation.
\end{itemize}


\section{Related Work}
\label{sec:related_work}
This section extends the comparisons with existing work presented in Section~\ref{intro}.

\subsection{Motion Prediction and Learned Traversability}
\label{sec:related_learning}

\noindent Traversability methods can assess predicted vehicle motion
or directly estimate terrain costs.
Takemura and Ishigami~\cite{takemura2022computationally}
combine a quasi-dynamic model with a tracking controller
to predict trajectories. They assess these trajectories
using limits on roll, pitch, and slip.
Learned predictive models also support motion-based
traversability assessment~\cite{kahn2020badgr,tal}.
Wang et al.~\cite{pmlr-v164-wang22c} combine learned
probabilistic dynamics with a tracking controller
to predict closed-loop executions through recursive
model--controller evaluations.

Other methods map terrain observations directly to
traversability costs or scores.
Castro et al.~\cite{guaman2023hdif} learn
velocity-conditioned costs from visual and geometric
features, using IMU measurements for supervision.
WVN~\cite{frey2023fast} predicts visual traversability
scores using velocity-tracking errors as self-supervision.
RoadRunner~\cite{frey2024roadrunner} predicts traversability
maps from camera and LiDAR measurements, using labels
generated in hindsight.
TNT~\cite{pan2024tnt} maps terrain elevation to
traversability costs. Its supervision uses vehicle
attitude, differences between commanded and executed
motion, and kinodynamic prediction uncertainty.
Other approaches learn terrain costs from expert
demonstrations through inverse reinforcement
learning~\cite{triest2023risk}, or directly predict
risk-sensitive costs~\cite{fan2022risk}.

\emph{Our FM model directly predicts a joint distribution
of complete closed-loop trajectories for a reference path
and a fixed vehicle--controller pair.
This avoids recursive dynamics--controller rollouts during
prediction. It also supports path-specific assessment
from predicted executions, rather than from local
terrain scores alone.}

\subsection{Probabilistic Traversability and Risk-Aware Planning}
\label{sec:related_probabilistic}

\noindent Probabilistic methods account for uncertainty
in traversability assessment.
Leininger et al.~\cite{leininger2024gaussian} constructed
sparse Gaussian-process-based elevation maps.
They used the derived terrain slope, uncertainty,
and footprint assessment as the traversability metric.
STEP~\cite{fan2021step} combines terrain uncertainty,
conditional value-at-risk (CVaR), and model predictive
control.
PIETRA~\cite{cai2025pietra} predicts uncertainty in traction
and vehicle attitude.
It also uses CVaR to assess traction risk and constrain
roll and pitch.

\emph{Our approach follows a risk-aware framework similar
to those of the above works. However, EGT provides the probabilistic
traversability estimate directly over closed-loop executions
for a given reference path and controller.
The effects of traction loss and tip-over are captured
when they cause violations of the tracking or progress
requirements.}


\section{Methods}\label{sec:method}
\noindent In this section, we first derive Execution-Grounded Traversability (EGT), assuming that a
conditional closed-loop predictor is available. We then describe a
latent-space flow-matching realization of this predictor and its use
in a receding-horizon, sampling-based planner.

\subsection{Execution-Grounded Traversability}
\label{sec:egt}

\noindent Let $\pi$ denote the fixed feedback controller. For simplicity, we consider
straight-line reference paths starting at the current vehicle
position $\mathbf{q}_0$. Reference waypoints are denoted by
$\mathbf{q}_k^{\mathrm{ff}}$ and are stacked to obtain a reference
$\boldsymbol{\tau}^{\mathrm{ff}}$ with the following arc-length parameterization:
\begin{equation}
    \boldsymbol{\gamma}(s) = \mathbf{q}_0+s\mathbf{d}, \qquad s\in[0,l], \qquad \|\mathbf{d}\|_2=1,
    \label{eq:straight_reference}
\end{equation}
where $\mathbf{d}\in\mathbb{R}^2$ is its direction and $l>0$ is its
length. Although the reference path is a straight line, the realized closed-loop execution can follow a curved trajectory determined by the vehicle's kinematics and the vehicle--terrain interaction dynamics. The vehicle starts tracking the reference path from rest and comes to a halt after the execution.

Let $\mathbf{C}$ be the terrain context formed by fusing $\boldsymbol{\tau}^{\mathrm{ff}}$, multi-resolution terrain features, and the vehicle's 3D pose prior. Let $\left(
            \mathbf{p}_0,\ldots,\mathbf{p}_h
        \right)$ be the sequence of waypoints obtained by tracking $\boldsymbol{\tau}^{\mathrm{ff}}$ using the controller $\pi$. The predicted execution waypoints
represent vehicle positions at $0.1\,\mathrm{s}$ intervals,
including any time spent at the endpoint after arrival. Assume that we have access to a conditional
closed-loop predictor, parameterized by $\boldsymbol{\theta}$, which
defines a distribution over complete execution trajectories:
\begin{equation}
    \boldsymbol{\tau}^{\mathrm{cl}} =\operatorname{col}(\mathbf{p}_0,\ldots,\mathbf{p}_h) \in\mathbb{R}^{2(h+1)},\quad \boldsymbol{\tau}^{\mathrm{cl}} \sim p_{\boldsymbol{\theta}}^\pi(\cdot\mid\mathbf{C}).
    \label{eq:execution_distribution}
\end{equation}
\noindent Here, $\operatorname{col}$ stacks its arguments into a column
vector. Here, $\mathbf{p}_k\in\mathbb{R}^2$ is the $k$th waypoint and
$h$ is the number of prediction steps. 

\paragraph{Tracking and reference-path progress}
Given a closed-loop trajectory $\boldsymbol{\tau}^{\mathrm{cl}}$, the arc-length
coordinate of the nearest reference point is obtained directly as
\begin{equation}
    s_k = \operatorname{clip}_{[0,l]}\!\left( \mathbf{d}^{\top}(\mathbf{p}_k-\mathbf{q}_0) \right), \qquad k=0,\ldots,h.
    \label{eq:path_projection}
\end{equation}
The dot product measures displacement along the reference direction,
while clipping restricts the projection to the reference segment.

Let $\mathbf{n}$ be a constant unit normal satisfying
$\mathbf{n}^{\top}\mathbf{d}=0$. The signed orthogonal tracking
error simplifies to
\begin{equation}
    e_{\perp,k} =\mathbf{n}^{\top}[\mathbf{p}_k-\boldsymbol{\gamma}(s_k)] =\mathbf{n}^{\top}(\mathbf{p}_k-\mathbf{q}_0).
    \label{eq:normal_error}
\end{equation}
For a prescribed corridor half-width $w(s)>0$, define the
normalized worst-case tracking error:
\begin{equation}
    z_\perp(\boldsymbol{\tau}^{\mathrm{cl}}) = \max_{k=0,\ldots,h} \frac{|e_{\perp,k}|}{w(s_k)}.
    \label{eq:tracking_statistic}
\end{equation}
The condition $z_\perp(\boldsymbol{\tau}^{\mathrm{cl}})\leq1$
requires the orthogonal tracking error to remain within the
prescribed tolerance at every waypoint of the predicted closed-loop trajectory.

\noindent We also define terminal reference-path progress as follows:
\begin{equation}
    \rho(\boldsymbol{\tau}^{\mathrm{cl}}) = \frac{s_h}{l} = \operatorname{clip}_{[0,1]}\!\left( \frac{\mathbf{d}^{\top}(\mathbf{p}_h-\mathbf{p}_0)}{l} \right).
    \label{eq:progress_statistic}
\end{equation}
This statistic measures advancement along the reference path $\boldsymbol{\tau}^{\mathrm{ff}}$. We use terminal rather than maximum attained progress to account for cases where the robot retreats after making some progress along the reference path. The progress requirement is
$\rho(\boldsymbol{\tau}^{\mathrm{cl}})\geq\beta$, where
$\beta\in(0,1]$ specifies the minimum completion fraction.

\begin{figure}[!t]
    \centering
    \includegraphics[width=0.45\textwidth]{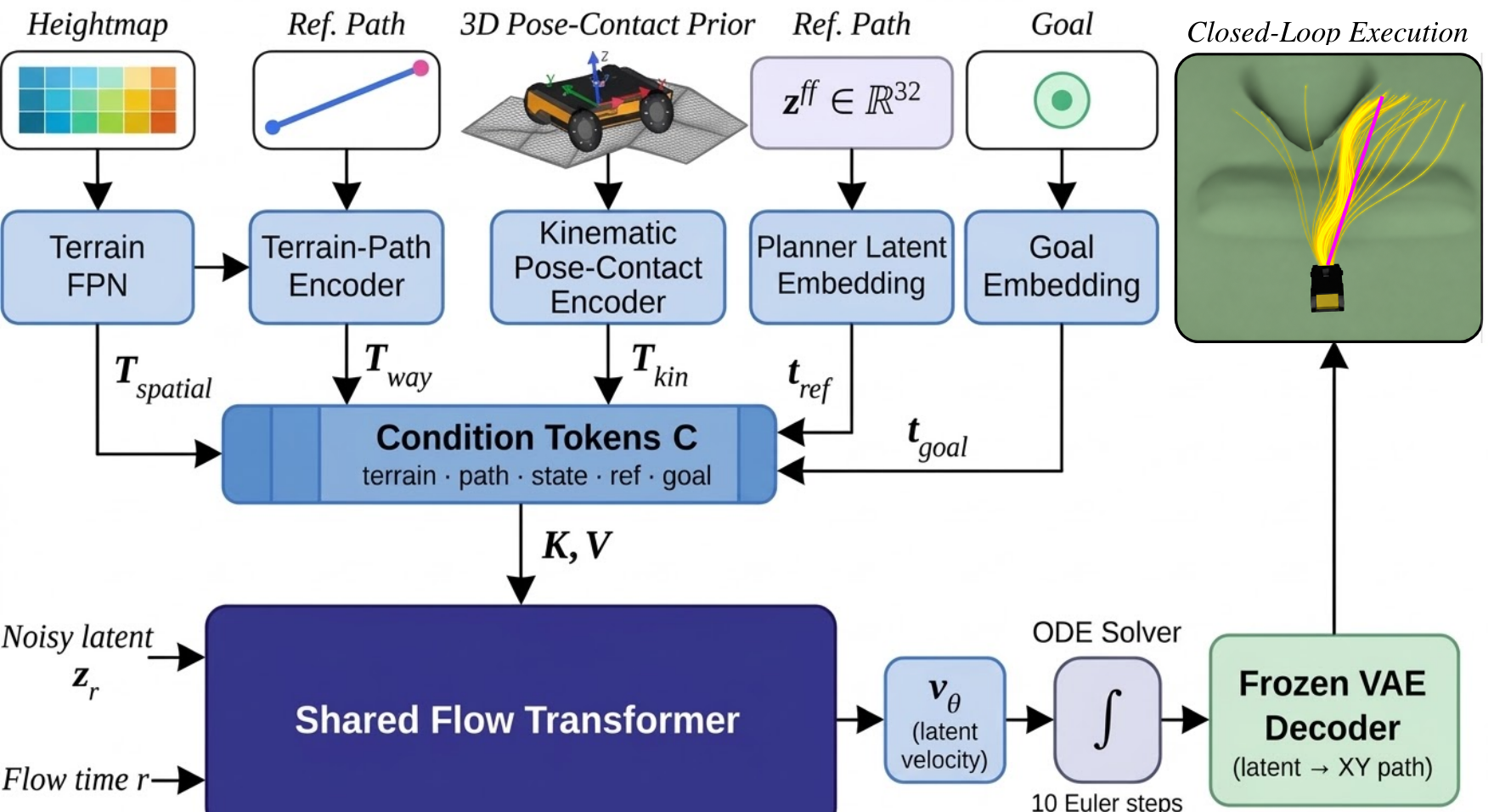}
    \caption{\footnotesize{Overview of the conditional latent flow matching architecture.
    Terrain FPN and path-fusion features, kinematic pose--contact features, and
    planner--goal context form conditioning
    tokens $\mathbf{C}$. A shared flow transformer predicts the conditional latent
    velocity field $v_\theta$, which is integrated and decoded into closed-loop
    trajectory rollouts.}}
    \label{fig:arc_overview}
\end{figure}

\paragraph{Distributional definition of EGT}
The conditional execution distribution induces a joint distribution
over the scalar quantities
$z_\perp(\boldsymbol{\tau}^{\mathrm{cl}})$ and
$\rho(\boldsymbol{\tau}^{\mathrm{cl}})$, denoted by $z_\perp$ and
$\rho$ for brevity. We define EGT as the probability that both
execution requirements hold simultaneously:
\begin{equation}
    \thinmuskip=2mu \medmuskip=2.5mu \thickmuskip=2.5mu
    p_{\mathrm{egt}}=\Pr_{\boldsymbol{\theta}}^\pi(z_\perp\leq1\land\rho\geq\beta\mid\mathbf{C})=\mathbb{E}_{\boldsymbol{\theta}}^\pi[\operatorname{ind}(z_\perp\leq1\land\rho\geq\beta)\mid\mathbf{C}].
    \label{eq:egt_probability}
\end{equation}
The probability and expectation are taken under the conditional
execution distribution in~\eqref{eq:execution_distribution}.
Here, $\operatorname{ind}(\cdot)$ equals one when its argument is
true and zero otherwise. Tracking accuracy and sufficient progress
must hold within the same execution.

\paragraph{Sample-average approximation}
To evaluate~\eqref{eq:egt_probability}, we draw $n_{\mathrm{s}}$
independent complete executions
$\widehat{\boldsymbol{\tau}}^{\mathrm{cl},(m)}
\sim p_{\boldsymbol{\theta}}^\pi(\cdot\mid\mathbf{C})$,
for $m=1,\ldots,n_{\mathrm{s}}$. For each execution, we compute
\begin{equation}
    z_\perp^{(m)} =z_\perp(\widehat{\boldsymbol{\tau}}^{\mathrm{cl},(m)}),\quad \rho^{(m)} =\rho(\widehat{\boldsymbol{\tau}}^{\mathrm{cl},(m)}).
    \label{eq:sample_statistics}
\end{equation}
The sample-average approximation of EGT is
\begin{equation}
    \widehat{p}_{\mathrm{egt}} = \frac{1}{n_{\mathrm{s}}} \sum_{m=1}^{n_{\mathrm{s}}} \operatorname{ind}\!\left( z_\perp^{(m)}\leq1 \;\land\; \rho^{(m)}\geq\beta \right).
    \label{eq:egt_estimate}
\end{equation}
\noindent Here, $\operatorname{ind}$ equals one if the condition holds and zero otherwise. This estimate is the fraction of predicted executions that jointly
satisfy the tracking and progress requirements.

\subsection{Learning Closed-Loop Execution Distributions}
\label{sec:fm}
\noindent In this section, we develop a latent-space FM model to approximate the distribution of closed-loop executions $p_{\boldsymbol{\theta}}^\pi$ for a given reference path and low-level controller $\pi$. Let $\boldsymbol{\tau}^{\mathrm{ff}}$ and
$\boldsymbol{\tau}^{\mathrm{cl}}_{gt}$ denote a reference path and its
corresponding ground-truth closed-loop execution. Both are represented
by $h+1=100$ ordered planar waypoints in the robot's local frame. Let
$\mathbf{e}_{\boldsymbol{\varphi}}$ and
$\mathbf{d}_{\boldsymbol{\varphi}}$ denote the variational autoencoder (VAE) encoding and
decoding operations, with parameters $\boldsymbol{\varphi}$.
The reference and execution codes are
\begin{equation}
    \mathbf{z}^{\mathrm{ff}} =\mathbf{e}_{\boldsymbol{\varphi}}(\boldsymbol{\tau}^{\mathrm{ff}}),\quad \mathbf{z}^{\mathrm{cl}} =\mathbf{e}_{\boldsymbol{\varphi}}(\boldsymbol{\tau}^{\mathrm{cl}}_{gt}),
    \label{eq:fm_vae_codes}
\end{equation}
where $\mathbf{z}^{\mathrm{ff}},\mathbf{z}^{\mathrm{cl}}
\in\mathbb{R}^{d_{\mathrm{z}}}$ and $d_{\mathrm{z}}=32$.
The VAE is trained beforehand, and both
its encoder and decoder remain frozen during FM learning. We use $\mathbf{z}^{\mathrm{ff}}$ as part of the conditioning input $\mathbf{C}$ in the FM model, along with additional tokens derived from terrain geometry and vehicle pose (see Fig.~\ref{fig:arc_overview} and Appendix~\ref{appendix}). The vector $\mathbf{z}^{\mathrm{cl}}$ is used as the training target.

\paragraph{Flow-matching training}
Each training example provides a conditioning
input and a paired ground-truth execution latent
$\mathbf{z}^{\mathrm{cl}}$. The aim of flow matching is to learn a velocity field that maps standard Gaussian noise to the distribution of closed-loop executions of the reference path. To this end, let
$\mathbf{z}_0\sim\mathcal{N}(\mathbf{0},\mathbf{I}_{d_{\mathrm{z}}})$
and $r\sim\mathcal{U}[0,1]$. Conditional flow
matching~\cite{lipman2023flowmatching} uses linear interpolation
and minimizes the following velocity-matching loss:
\begin{align}
    \mathbf{z}_r &=(1-r)\mathbf{z}_0+r\mathbf{z}^{\mathrm{cl}},
    \label{eq:fm_latent_interpolation}\\
    \ell_{\mathrm{flow}} &=\mathbb{E}\!\left[ \left\|\mathbf{v}_{\boldsymbol{\theta}}(\mathbf{z}_r,r,\mathbf{C}) -(\mathbf{z}^{\mathrm{cl}}-\mathbf{z}_0)\right\|_2^2 \right],
    \label{eq:fm_velocity_loss}
\end{align}
where the expectation is over training examples, Gaussian noise,
and flow time $r$. The velocity field $\mathbf{v}_{\boldsymbol{\theta}}$ is modeled as a sequence of transformer blocks in our implementation, with learnable weights $\boldsymbol{\theta}$.

We also include auxiliary losses to supervise the decoded waypoint
coordinates and terminal position. During training, we unroll the
learned latent dynamics and decode their endpoint:
\begin{align}
    \frac{d\widehat{\mathbf{z}}(r)}{dr} &=\mathbf{v}_{\boldsymbol{\theta}} (\widehat{\mathbf{z}}(r),r,\mathbf{C}),
    \label{eq:fm_latent_ode}\\
    \widehat{\boldsymbol{\tau}}^{\mathrm{cl}} &=\mathbf{d}_{\boldsymbol{\varphi}}(\widehat{\mathbf{z}}^{\mathrm{cl}}) =\operatorname{col}(\widehat{\mathbf{p}}_0,\ldots,\widehat{\mathbf{p}}_h),
    \label{eq:fm_training_decode}
\end{align}
where $\widehat{\mathbf{z}}^{\mathrm{cl}}=\widehat{\mathbf{z}}(1)$
and $\operatorname{col}$ represents column-wise stacking.
Writing $\mathbf{p}_k^{\mathrm{cl}}$ for the corresponding
ground-truth execution waypoint, the auxiliary losses and total
training objective are
\begin{align}
    \ell_{\mathrm{coord}}&\!=\!\mathbb{E}\!\left[\tfrac{1}{h+1}\!\sum_{k=0}^{h}\!\|\widehat{\mathbf{p}}_k\!-\!\mathbf{p}_k^{\mathrm{cl}}\|_2^2\right],\;
    \ell_{\mathrm{end}}\!=\!\mathbb{E}\!\left[\|\widehat{\mathbf{p}}_h\!-\!\mathbf{p}_h^{\mathrm{cl}}\|_2^2\right],
    \label{eq:fm_auxiliary_losses}\\
    \ell_{\mathrm{total}}&=\ell_{\mathrm{flow}}+\lambda_{\mathrm{coord}}\ell_{\mathrm{coord}}+\lambda_{\mathrm{end}}\ell_{\mathrm{end}},
    \label{eq:fm_total_loss}
\end{align}
with $\lambda_{\mathrm{coord}}=2$ and $\lambda_{\mathrm{end}}=4$ chosen for our implementation.
Gradients from the auxiliary losses pass through the decoder to
the flow model, while the VAE parameters remain fixed.
We use AdamW with an initial learning rate of $10^{-4}$,
$10$ warmup epochs, a cosine learning-rate schedule over $500$
epochs, and an exponential moving average with decay $0.999$.

\paragraph{Sampling complete executions}
At inference time, the conditioning matrix $\mathbf{C}$ is constructed
once for each reference and reused across flow evaluations and
execution samples. For each sample, we initialize
$\widehat{\mathbf{z}}(0)\sim
\mathcal{N}(\mathbf{0},\mathbf{I}_{d_{\mathrm{z}}})$ and integrate
\eqref{eq:fm_latent_ode} using $n_{\mathrm{flow}}=10$ Euler steps:
\begin{equation}
    \widehat{\mathbf{z}}(r_{j+1}) =\widehat{\mathbf{z}}(r_j)+\delta_r\, \mathbf{v}_{\boldsymbol{\theta}} (\widehat{\mathbf{z}}(r_j),r_j,\mathbf{C}),
    \label{eq:fm_euler_sampling}
\end{equation}
where $r_j=j\delta_r$, $\delta_r=1/n_{\mathrm{flow}}$, and
$j=0,\ldots,n_{\mathrm{flow}}-1$. The final latent is decoded into the full waypoint sequence.
Before evaluating EGT, we smooth the decoded sequence using Bernstein polynomials and
shift it so that its initial waypoint coincides with the local origin. Repeating this procedure with independent noise draws provides
the execution samples used in~\eqref{eq:egt_estimate}.

\paragraph{Overview of the neural architecture and structure of the conditioning input} The conditioning input $\mathbf{C}$ has the following structure:
\begin{equation}
    \mathbf{C} =\begin{bmatrix} \mathbf{t}_{\mathrm{ref}} & \mathbf{t}_{\mathrm{goal}} & \mathbf{T}_{\mathrm{way}}^{\top} & \mathbf{T}_{\mathrm{kin}}^{\top} & \mathbf{T}_{\mathrm{spatial}}^{\top} \end{bmatrix}^{\top},
    \label{eq:fm_condition_memory}
\end{equation}
where $\mathbf{C}\in\mathbb{R}^{n_{\mathrm{mem}}\times d}$ and
$n_{\mathrm{mem}}=2+2(h+1)+n_3=266$. We use a feature pyramid network
(FPN)~\cite{lin2017feature} to build the tokens
$\mathbf{T}_{\mathrm{spatial}}$ and $\mathbf{T}_{\mathrm{way}}$.
These tokens capture a coarse overview of the terrain and
fine-grained terrain features near the reference path, respectively.
Similarly, $\mathbf{t}_{\mathrm{ref}}$, $\mathbf{T}_{\mathrm{kin}}$,
and $\mathbf{t}_{\mathrm{goal}}$ are formed by embedding,
respectively, the latent representation of the reference path,
the vehicle's 3D pose and contact points obtained using the method
in~\cite{manoharan2024bi}, and the goal or endpoint of the reference
path. Appendix~\ref{appendix} gives further details of the token
construction and the latent-space transformer block.


\subsection{Risk-Aware Sampling-Based Planning}
\label{sec:planner}

\noindent We use the developments of the previous subsections to design a sampling-based receding-horizon planner for rough-terrain navigation. At each replanning cycle, we generate a set of $n_{\mathrm{c}}$
candidate references,
$\mathcal{C}=\{\boldsymbol{\tau}_i^{\mathrm{ff}}\}_{i=1}^{n_{\mathrm{c}}}$.
We construct the context $\mathbf{C}_i$ for each candidate path and subsequently sample
$n_{\mathrm{s}}$ executions
$\widehat{\boldsymbol{\tau}}_i^{\mathrm{cl},(m)}$ from the learned FM model. The superscript identifies the closed-loop execution sample, while the subscript identifies the candidate path. Applying the procedure in Section~\ref{sec:egt} to the sampled closed-loop executions gives $z_{\perp,i}^{(m)}$,
$\rho_i^{(m)}$, and $\widehat{p}_{\mathrm{egt},i}$.

We adopt a two-step process for selecting the best candidate.
We first construct the EGT-accepted safe set $\mathcal{F}$ and then rank its
members using the cost $j_i$:
\begin{align}
    \mathcal{F} &=\left\{i\in\{1,\ldots,n_{\mathrm{c}}\}: \widehat{p}_{\mathrm{egt},i}\geq p_{\mathrm{req}}\right\},
    \label{eq:accepted_candidates}\\
    j_i &=\lambda_{\mathrm{g}}\widetilde{c}_{\mathrm{g},i} +\lambda_\perp r_{\perp,i} +\lambda_{\mathrm{prog}}r_{\mathrm{prog},i}.
    \label{eq:normal_objective}
\end{align}
The set $\mathcal{F}$ contains all candidate paths whose EGT
meets the specified threshold $p_{\mathrm{req}}$.
\noindent The first term measures progress toward the global goal, while the last
two terms are CVaR estimates of the tracking loss $z_{\perp,i}$ and
progress loss $1-\rho_i$. These terms regulate worst-case path-tracking
and progress behavior. The least-cost path is then selected for execution.

\begin{figure}[!t]
    \centering
    \begin{subfigure}[t]{\columnwidth}
        \centering
        \includegraphics[width=0.32\linewidth]{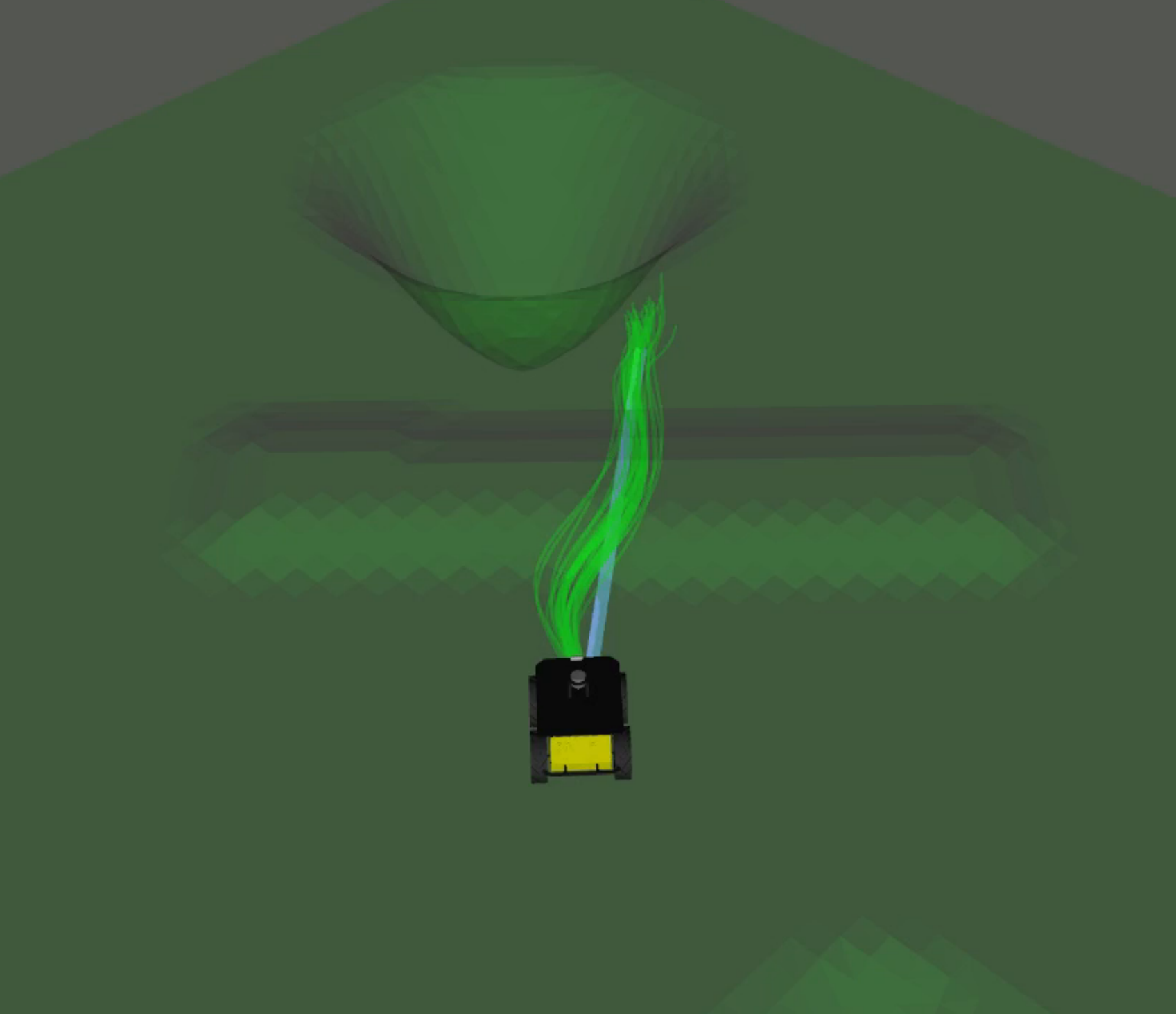}\hfill
        \includegraphics[width=0.32\linewidth]{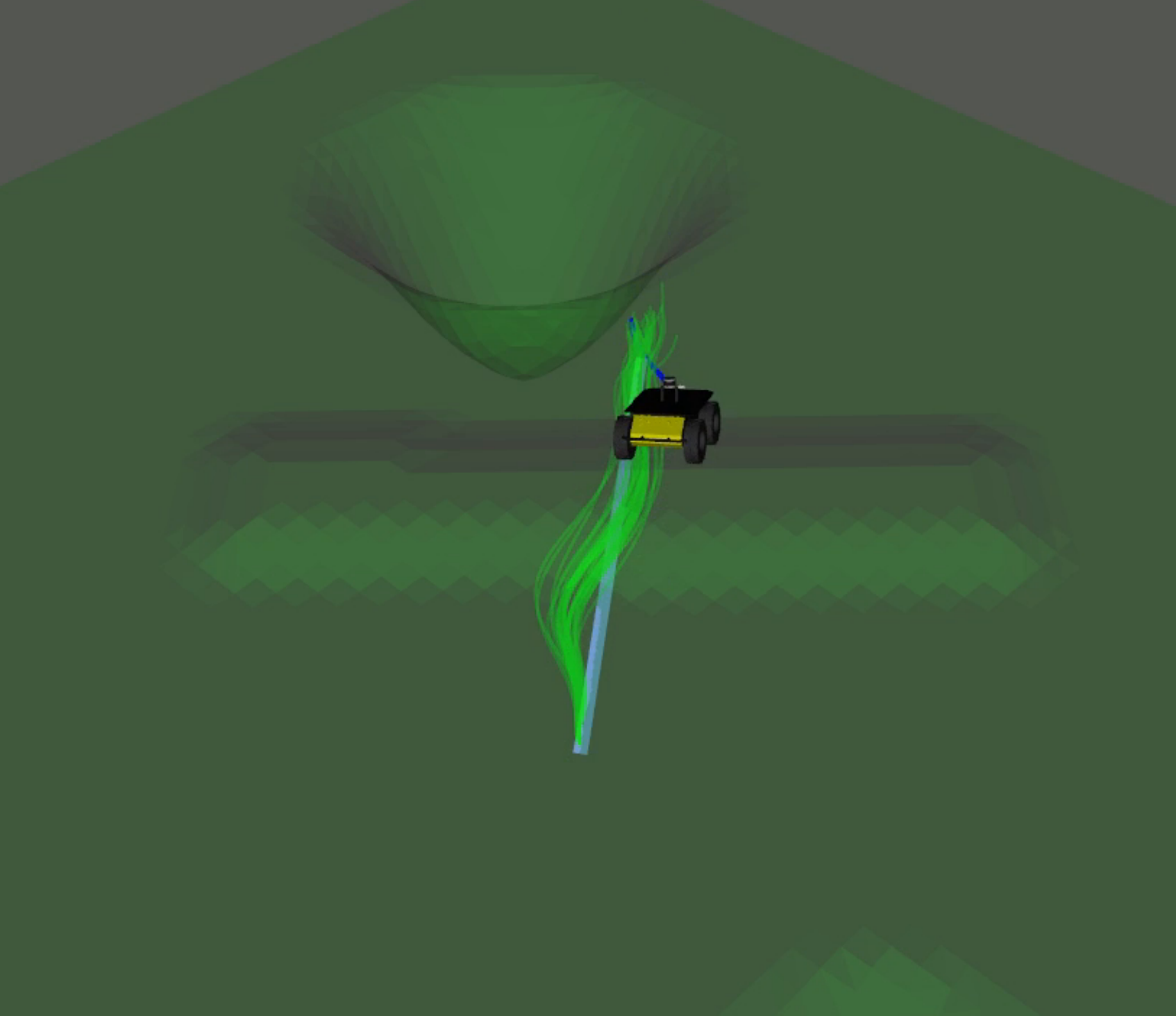}\hfill
        \includegraphics[width=0.32\linewidth]{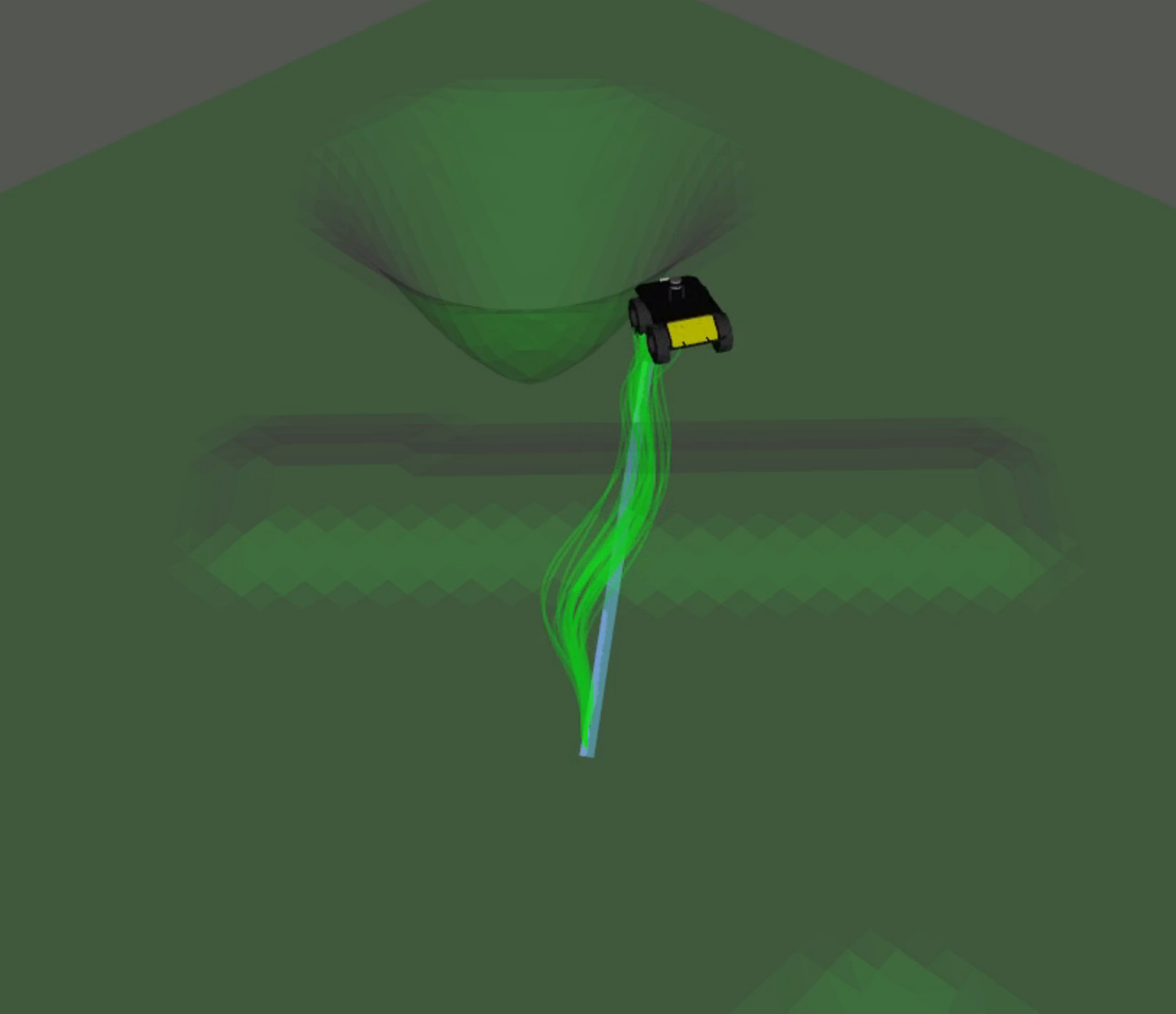}
        \caption{\footnotesize The FM closed-loop predictions (green) and the resulting EGT assessment indicate that the vehicle can clear the bump. The Gazebo simulation confirms this assessment.}
    \end{subfigure}
    \vspace{2pt}

    \begin{subfigure}[t]{\columnwidth}
        \centering
        \includegraphics[width=0.32\linewidth]{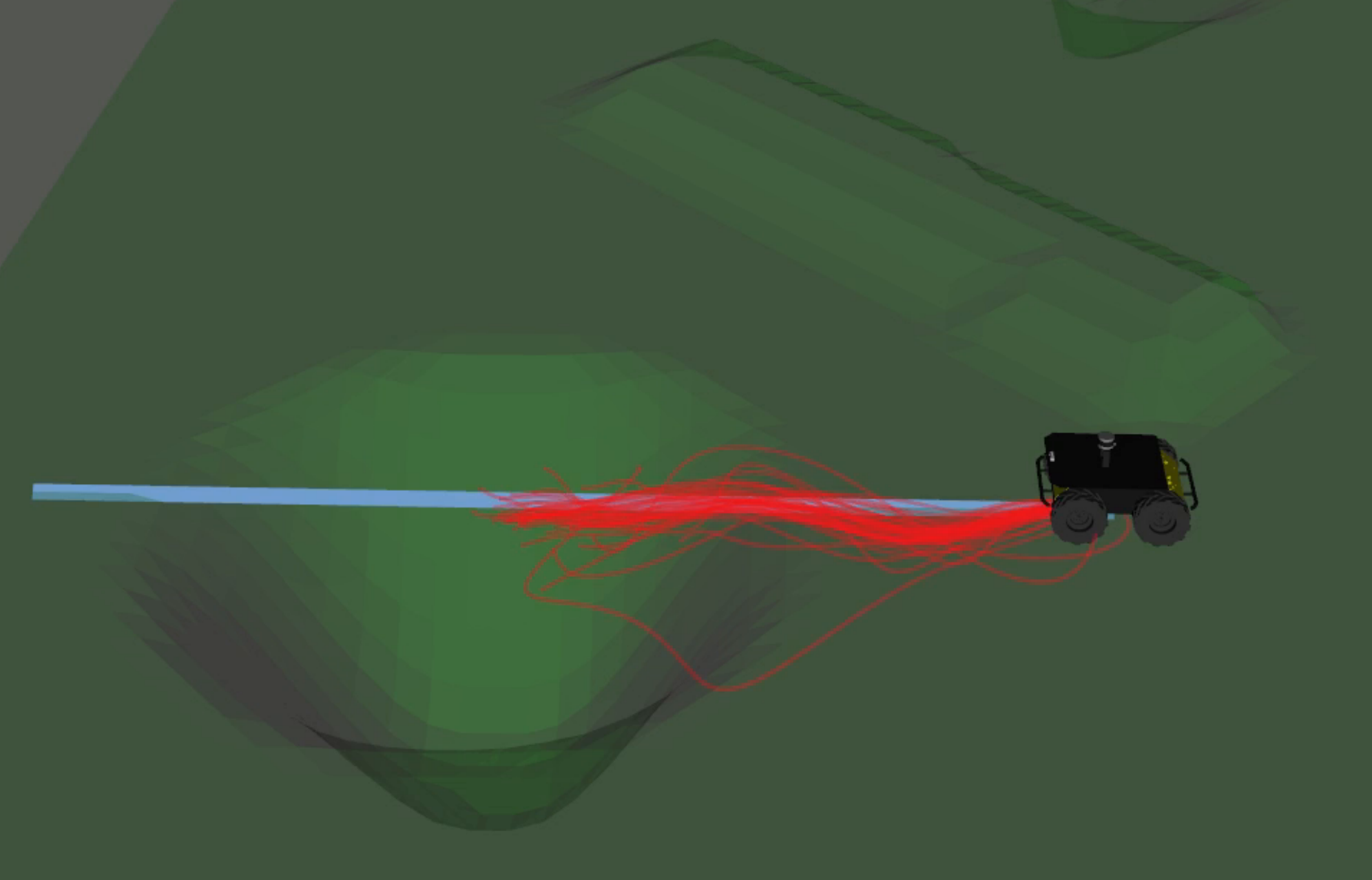}\hfill
        \includegraphics[width=0.32\linewidth]{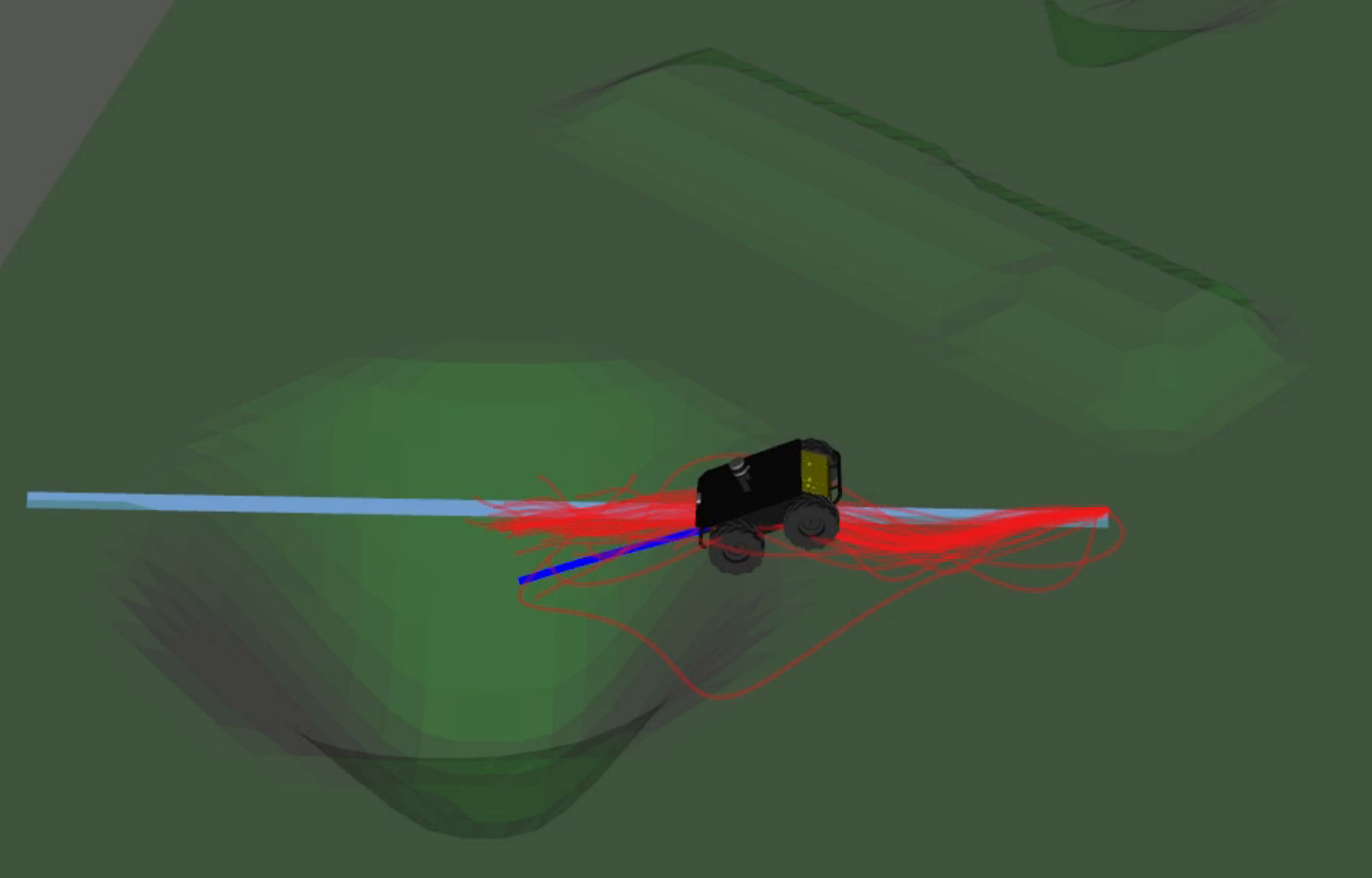}\hfill
        \includegraphics[width=0.32\linewidth]{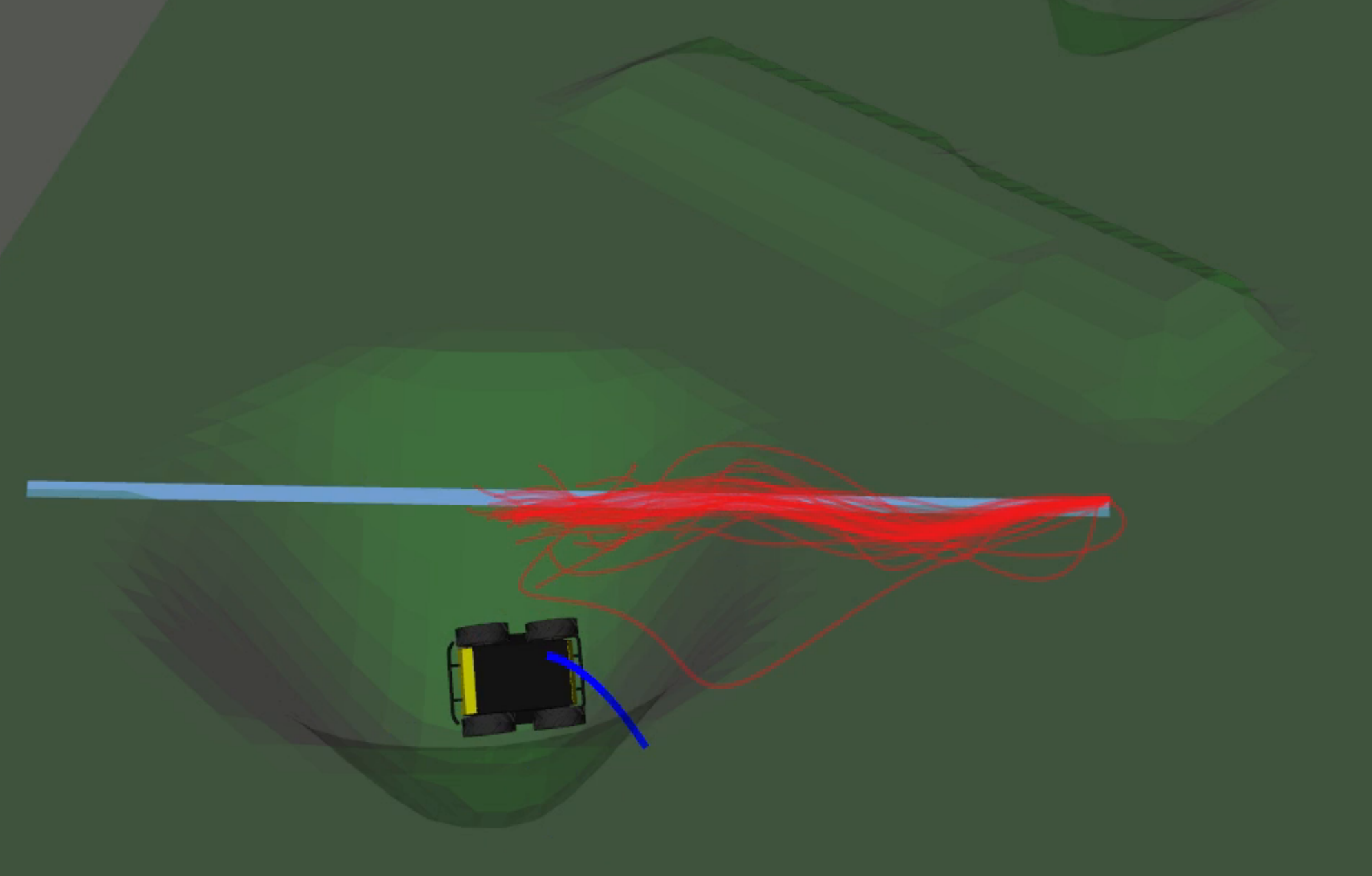}
        \caption{\footnotesize The FM closed-loop execution predictions (red) for a straight path over a ditch lead to a non-traversable EGT assessment. The Gazebo simulations confirm that the vehicle topples while crossing the ditch. The vehicle's terminal position is close to where the FM predictions end.}
    \end{subfigure}
    \vspace{2pt}

    \begin{subfigure}[t]{\columnwidth}
        \centering
        \includegraphics[width=0.32\linewidth]{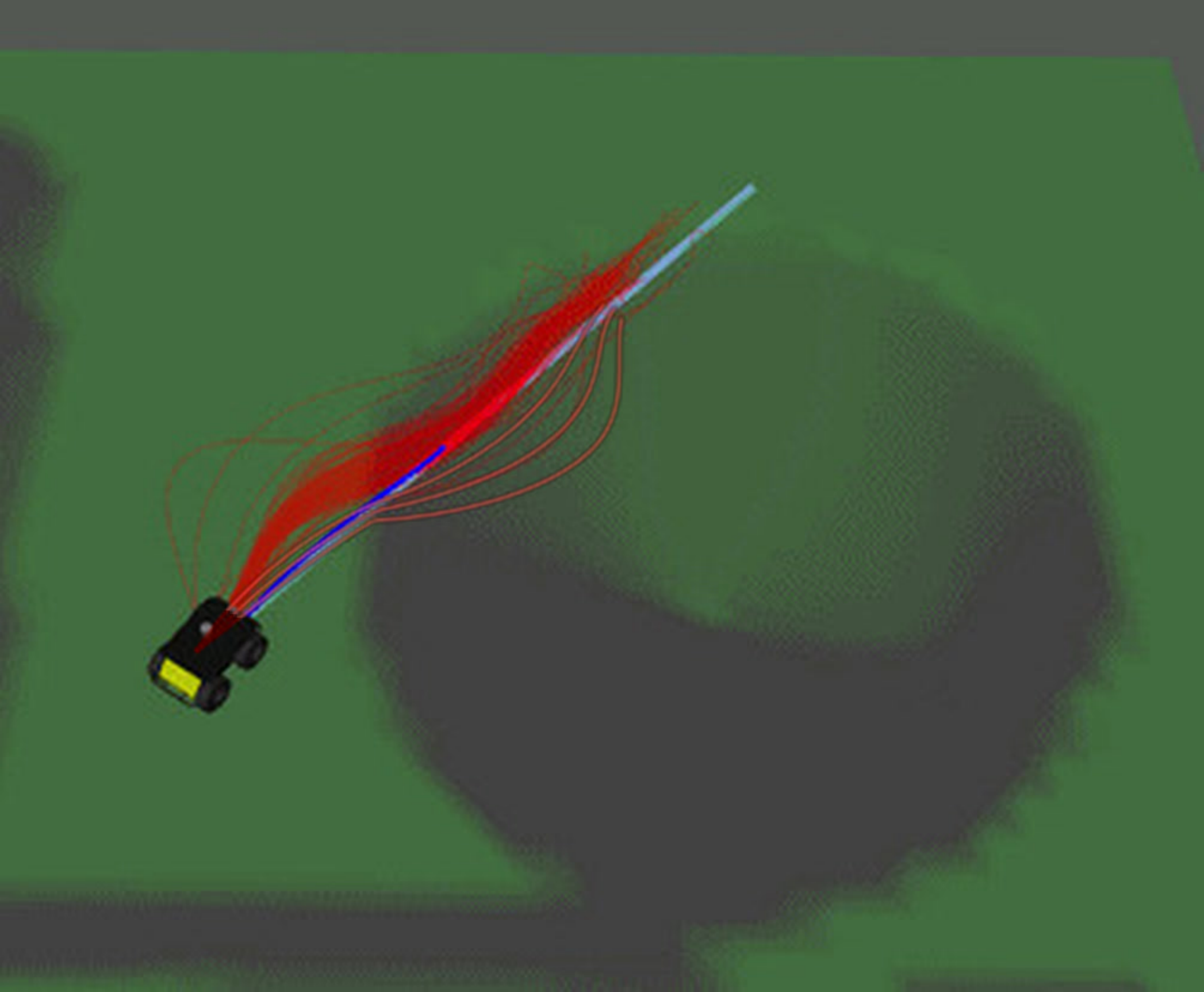}\hfill
        \includegraphics[width=0.32\linewidth]{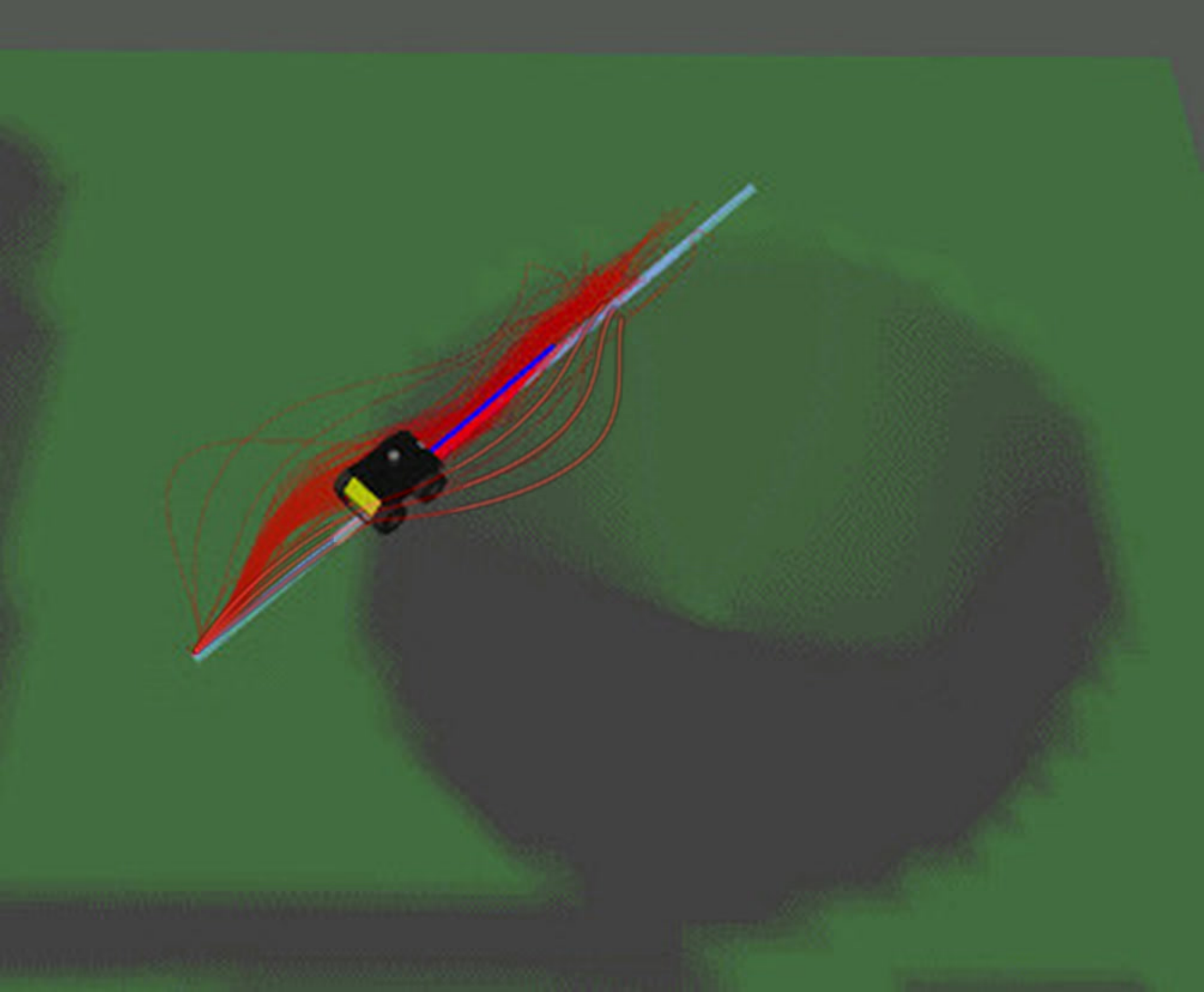}\hfill
        \includegraphics[width=0.32\linewidth]{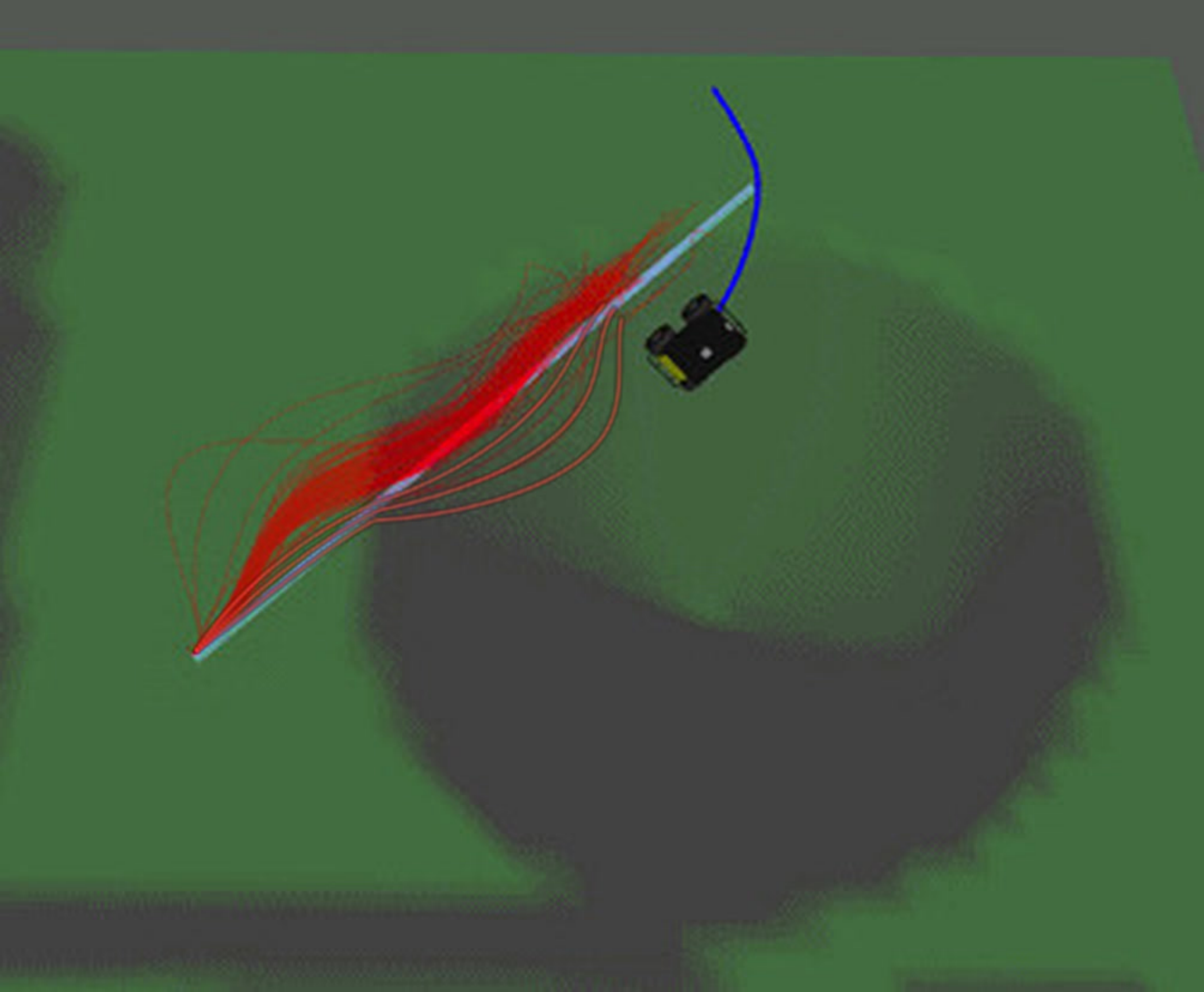}
        \caption{\footnotesize A scenario in which the reference path passes partly through a crater. The FM predictions (red) and the resulting EGT assessment indicate navigation failure due to uncertainty in tracking the path closely. The simulation run confirms this assessment.}
    \end{subfigure}
    \caption{\footnotesize{Representative closed-loop trajectory predictions on simulated terrain.}}
    \label{fig:simulation_predictions}
\end{figure}

\paragraph{Global task cost and execution risk}
Let $v_{\mathrm{g}}(\mathbf{p})$ denote the global cost-to-go from
position $\mathbf{p}$ to the mission goal $\mathbf{p}_{\mathrm{g}}$.
In open terrain, it can be the Euclidean distance
$\|\mathbf{p}-\mathbf{p}_{\mathrm{g}}\|_2$; when obstacles or route
topology matter, a map-aware cost-to-go can be used instead. The expected
terminal task cost and its normalized form are
\begin{equation}
    c_{\mathrm{g},i} =\frac{1}{n_{\mathrm{s}}}\sum_{m=1}^{n_{\mathrm{s}}} v_{\mathrm{g}}(\widehat{\mathbf{p}}_{i,h}^{(m)}),\quad \widetilde{c}_{\mathrm{g},i} =\frac{c_{\mathrm{g},i}}{v_{\mathrm{g}}(\mathbf{p}_0)+\epsilon},
    \label{eq:global_cost}
\end{equation}
where $\epsilon>0$. This term evaluates global task utility
separately from local reference-path progress.

We quantify adverse execution outcomes using empirical CVaR estimates. For scalar loss samples
$z^{(1)},\ldots,z^{(n_{\mathrm{s}})}$ and level $\alpha\in[0,1)$, we define
\begin{equation}
    \widehat{\operatorname{CVaR}}_\alpha(z) =\min_{\eta\in\mathbb{R}} \left\{ \eta+ \frac{\sum_{m=1}^{n_{\mathrm{s}}}[z^{(m)}-\eta]_+} {n_{\mathrm{s}}(1-\alpha)} \right\},
    \label{eq:empirical_cvar}
\end{equation}
where $[a]_+=\max(a,0)$. The tracking and progress risks are
\begin{equation}
    r_{\perp,i} =\widehat{\operatorname{CVaR}}_{\alpha_\perp}(z_{\perp,i}),\; r_{\mathrm{prog},i} =\widehat{\operatorname{CVaR}}_{\alpha_{\mathrm{prog}}}(1-\rho_i).
    \label{eq:execution_risks}
\end{equation}
The first term emphasizes large tracking deviations, while the
second emphasizes the low-progress tail.

\paragraph{Fallback when no candidate is accepted}

If no candidate meets $p_{\mathrm{req}}$, we relax the threshold to
within $\delta_p$ of the maximum candidate reliability, with
$p_{\mathrm{abort}}$ as the lower bound. We then repeat candidate
selection using the objective $j_i$ and the cached evaluations.
If no candidate meets $p_{\mathrm{abort}}$, the planner returns
no-go and regenerates paths. This fallback reduces the required
execution confidence rather than relaxing the tracking or
progress tolerances.

\section{Validation and Benchmarking}
This section has two objectives. First, we evaluate how accurately EGT captures a wheeled vehicle's ability to traverse a given rough terrain. Second, we assess the impact of EGT on downstream planning performance by benchmarking against existing navigation algorithms. 

\subsection{Implementation Details}
\noindent We implemented the FM model, EGT assessment, and sampling-based
planner in JAX, using Equinox~\cite{kidger2021equinox} as the training
framework. We adapt the method in~\cite{pimppi} as our low-level
path-tracking controller. The training data were collected in
ROS Gazebo using a Husky vehicle. Intel RealSense was used to
obtain terrain information in both simulation and real-world runs.
The training dataset consisted of closed-loop executions of
8,808 feed-forward trajectories on 7 different terrains. 

For all experiments, we use a constant corridor half-width
$w(s)=1.0\,\mathrm{m}$, minimum progress fraction $\beta=0.8$, and
required EGT probability $p_{\mathrm{req}}=0.8$. Each candidate is
evaluated using $n_{\mathrm{s}}=30$ predicted executions. At every
replanning cycle, the planner evaluates $n_{\mathrm{c}}=100$
candidate reference paths. Both execution-risk terms use a CVaR
level of $\alpha_{\perp}=\alpha_{\mathrm{prog}}=0.90$, and the
objective weights in~\eqref{eq:normal_objective} are
$\lambda_{\mathrm{g}}=\lambda_{\perp}=\lambda_{\mathrm{prog}}=0.5$.

Real-world implementation was done on a skid-steered robot Jackal which is similar to the vehicle used in training. The computations were performed on the mounted RTX 3090 laptop. The planning took an average of 1s, while the low-level path-tracking controller ran at 20Hz. 

The vehicle remains at rest during replanning and begins
tracking the selected reference only after planning
is complete. Each new execution therefore starts from
rest, consistent with the initial-state assumption
of the FM predictor.

\subsection{Qualitative Demonstration of EGT}
\label{sec:qualitative_egt}

\noindent Fig.~\ref{fig:simulation_predictions} provides a qualitative
validation of the FM-derived EGT assessment across three terrains using
$\widehat{p}_{\mathrm{egt}}$ from \eqref{eq:egt_estimate} (with $p_{req} = 0.80$).
In Fig.~\ref{fig:simulation_predictions}(a), the reference path crosses a
bump that the vehicle can traverse; the green FM predictions closely
follow the path to the goal, satisfying $\widehat{p}_{\mathrm{egt}} \geq p_{req}$
and classifying the path as traversable, as confirmed in Gazebo.
Conversely, in Fig.~\ref{fig:simulation_predictions}(b), the path crosses a
steep ditch and the predicted executions terminate early (recall
\eqref{eq:progress_statistic} and \eqref{eq:egt_probability}), yielding
$\widehat{p}_{\mathrm{egt}} < p_{req}$ (non-traversable). Simulation confirms
that the vehicle topples near the predicted termination point.
In Fig.~\ref{fig:simulation_predictions}(c), the predicted executions
exhibit large variance; although some samples succeed,
$\widehat{p}_{\mathrm{egt}}$ falls below $p_{req}$ and the path is classified
as non-traversable, matching repeated simulations in which the vehicle
falls in more frequently than it succeeds.

Fig.~\ref{fig:teaser_overview} shows a real-world validation of the EGT
assessment, where reference paths passing through non-navigable barriers
(lamppost in Trial B, hanging tree branch in Trial C) are correctly
classified as non-traversable.

\subsection{Quantitative Benchmarking of EGT}
\label{sec:quantitative_egt}

\noindent EGT classifies a path as traversable when
$\widehat{p}_{\mathrm{egt}}\geq p_{\mathrm{req}}$,
with $p_{\mathrm{req}}=0.8$ and computed from the predictions of the closed-loop executions through FM.
We compare it with two alternatives:

\begin{itemize}[leftmargin=*]
   \item \textbf{Force-Angle
Measure~\cite{Papadopoulos01012000}:}
This baseline evaluates tip-over stability along the
reference path using gravitational loading only.
We use~\cite{manoharan2024bi} to estimate the vehicle's
3D pose and wheel--ground contact points.
At each evaluated reference waypoint, we compute the
force-angle margins about all tip-over axes.
A path is accepted only if the minimum margin is
strictly greater than zero at every evaluated waypoint.
Inertial loads are not included in this baseline.

    \item \textbf{Pose-Margin:}
    Based on~\cite{takemura2022computationally,krusi2017driving},
    this baseline checks whether the vehicle's pitch
    and roll angles along the reference path exceed
    a set threshold. We evaluate thresholds of
    $45^\circ$, $60^\circ$, and $75^\circ$ to examine
    the trade-off between conservative rejection
    and incorrect acceptance.
\end{itemize}

\noindent\textbf{Methodology:}
We evaluate traversability classification over
650 simulated trials. All trials use terrains that
were not included in the FM training data.
In each trial, the vehicle--controller pair executes
a reference path once in simulation.
An execution is labeled safe if the vehicle reaches
the goal without tipping over, getting stuck, or
exceeding the prescribed lateral-deviation limit.
Otherwise, it is labeled unsafe.
The lateral limit is the same as the tracking tolerance
used in EGT, $w(s)=1.0\,\mathrm{m}$.
Here, safe and unsafe refer to successful and
unsuccessful reference-path execution under these
task requirements.
All methods are evaluated against the same outcome
labels, which are assigned from the actual recorded simulation.

The dataset contains $\overline{n}_{\mathrm{s}}=598$
safe and $\overline{n}_{\mathrm{u}}=52$ unsafe executions.
Let $\overline{n}_{\mathrm{r}}$ denote the number of
safe executions incorrectly rejected, and
$\overline{n}_{\mathrm{a}}$ the number of unsafe
executions incorrectly accepted.
The \emph{conservative rejection rate},
$100\,\overline{n}_{\mathrm{r}}/\overline{n}_{\mathrm{s}}$,
measures how often a method rejects paths that the
vehicle--controller pair successfully executes.
The \emph{unsafe acceptance rate},
$100\,\overline{n}_{\mathrm{a}}/\overline{n}_{\mathrm{u}}$,
measures how often a method accepts paths whose
recorded executions fail.
These error rates use their respective ground-truth
class sizes, rather than the total number of trials.
Finally, \emph{overall accuracy} is the percentage
of correctly classified trials, calculated as
$100\left[1-
(\overline{n}_{\mathrm{r}}+\overline{n}_{\mathrm{a}})
/
(\overline{n}_{\mathrm{s}}+\overline{n}_{\mathrm{u}})
\right]$.
Lower values are preferable for both error rates,
whereas higher accuracy is preferable.

\begin{table}[!t]
    \centering
    \caption{\footnotesize{Traversability classification
    over 650 simulated trials: 598 safe and 52 unsafe
    executions under the stated task requirements.
    All metric values are percentages.
    The EGT ablation omits conditioning on the NLS
    pose prior. Arrows indicate the preferred direction,
    and best values are shown in bold. The force-angle baseline uses gravitational loading only.}}
    \label{tab:egt_classification}
    \footnotesize
    \setlength{\tabcolsep}{3pt}
    \renewcommand{\arraystretch}{1.05}
    \resizebox{\columnwidth}{!}{%
    \begin{tabular}{@{}lcccccc@{}}
        \toprule
        Metric & Force-Angle \cite{Papadopoulos01012000}
        & \multicolumn{3}{c}{Pose-Margin}
        & \multicolumn{2}{c}{EGT} \\
        \cmidrule(lr){3-5}
        \cmidrule(lr){6-7}
        & & $45^\circ$ & $60^\circ$ & $75^\circ$
        & (Ours) & \shortstack{w/o pose-\\contact prior} \\
        \midrule
        Conservative rejection $\downarrow$
        & 39.5 & 35.3 & 21.7 & 11.4
        & \textbf{2.2} & 4.8 \\
        Unsafe acceptance $\downarrow$
        & 13.5 & 13.5 & 38.5 & 63.5
        & \textbf{7.7} & 71.2 \\
        Overall accuracy $\uparrow$
        & 62.6 & 66.5 & 76.9 & 84.5
        & \textbf{97.4} & 89.8 \\
        \bottomrule
    \end{tabular}%
    }
\end{table}

\noindent\textbf{Results:}
Table~\ref{tab:egt_classification} shows that the
evaluated tip-over and pose-margin criteria can be
both conservative and inaccurate as traversability
predictors. The force-angle baseline rejects $39.5\%$
of successful executions, yet accepts $13.5\%$ of
unsuccessful executions.
The pose-margin baselines reject
$11.4\%$--$35.3\%$ of successful executions and
accept $13.5\%$--$63.5\%$ of unsuccessful executions.
Increasing the pose-margin threshold reduces
conservative rejection but increases incorrect
acceptance.

EGT achieves the lowest observed rates of both errors
and the highest overall accuracy.
It incorrectly rejects 13 of 598 successful executions
and accepts four of 52 unsuccessful executions.
These counts give a conservative-rejection rate of
$2.2\%$, an unsafe-acceptance rate of $7.7\%$,
and an overall classification accuracy of $97.4\%$.

Without conditioning on the 3D pose and contact-point prior from \cite{manoharan2024bi}, EGT
rejects 29 of 598 successful executions and accepts
37 of 52 unsuccessful executions.
The corresponding error rates increase to $4.8\%$
and $71.2\%$, while overall accuracy decreases
to $89.8\%$. This ablation shows the importance
of the pose prior for the evaluated execution predictor. It should be noted that for the pose-prior ablation, we train a separate FM model
without pose--contact conditioning, using the same
training dataset and training procedure.

Overall, these results provide two key validations. First, the pose limits and
tip-over stability margins measured through force-angle measure, can reject executable
paths while also often accepting paths that fail the execution
requirements. Second, EGT probability estimated through the learned FM model reduces both types of error by
assessing predicted closed-loop execution.


\subsection{Probability Calibration}
\noindent We use the same $N=650$ simulated path executions
and outcome labels as Table~\ref{tab:egt_classification}
to assess the calibration of EGT as a probability forecast
of successful path execution.
Let $p_i=\widehat p_{\mathrm{egt},i}$ for the $i$th
reference path. We set $y_i=1$ if the recorded execution
meets the success criteria in
Section~\ref{sec:quantitative_egt}, and $y_i=0$ otherwise.
To recall, success requires reaching the goal without getting stuck, tipping over or violating the prescribed lateral limit.
The outcome labels $y_i$, are obtained from actual simulated motion. In contrast, $p_i$ is computed from the predictions from the FM model.

We evaluate calibration by comparing the predicted
probabilities with the observed execution-success
frequencies. Each reference path is executed only once.
Therefore, we group the test cases into probability bins
to compare the predictions with multiple recorded outcomes.
We use bins of unequal width because uniform division
resulted in some bins with too few cases.
Let $\mathcal{B}_b$ be the set of cases in bin $b$,
with $n_b=|\mathcal{B}_b|$.
For each bin, we calculate the mean predicted probability
$\bar p_b=n_b^{-1}\sum_{i\in\mathcal{B}_b}p_i$
and the observed success frequency
$\bar y_b=n_b^{-1}\sum_{i\in\mathcal{B}_b}y_i$.
Here, $\bar y_b$ is the fraction of recorded executions
in bin $b$ that meet the stated success criteria.
Agreement between $\bar p_b$ and $\bar y_b$ indicates
calibration at the group level.
We report pointwise $95\%$ Wilson confidence
intervals~\cite{wilson1927probable} for the group
execution-success probabilities to describe uncertainty
due to the finite number of recorded outcomes.

Table~\ref{tab:egt_calibration} shows that EGT
underestimates the success frequency in the lowest bin, wherein
the mean prediction is $0.0656$, whereas the observed
frequency is $0.2131$.
The confidence interval is $[0.1290,0.3312]$,
which is entirely above the mean prediction.
In the two intermediate bins, all recorded executions
meet the success criteria, and the mean predictions
are within the confidence intervals.
For predictions where $p_i=1$, 481 of the 485
executions meet these criteria. In other words, EGT predicts perfect success against a 
the observed success frequency of $0.9918$.

To summarize, Table~\ref{tab:egt_calibration} shows calibration varies across the bins: the predictions
are conservative in the lower bins, but they slightly
overestimate the success frequency when $p_i=1$.



\begin{table}[!t]
    \centering
    \caption{\footnotesize{EGT probability calibration
    against the execution-success labels used in
    Table~\ref{tab:egt_classification}.
    $n_b$: bin size;
    $\bar p_b$: mean predicted probability;
    $\bar y_b$: observed execution-success frequency;
    CI: pointwise $95\%$ Wilson confidence interval
    for the group execution-success probability.}}
    \label{tab:egt_calibration}
    \scriptsize
    \setlength{\tabcolsep}{3pt}
    \begin{tabular}{@{}crrrc@{}}
        \hline
        Probability bin & $n_b$ & $\bar p_b$
        & $\bar y_b$ & $95\%$ CI \\
        \hline
        $[0,0.80)$
        & 61 & 0.0656 & 0.2131
        & $[0.1290,0.3312]$ \\
        $[0.80,0.96)$
        & 40 & 0.9200 & 1.0000
        & $[0.9124,1.0000]$ \\
        $[0.96,1)$
        & 64 & 0.9667 & 1.0000
        & $[0.9434,1.0000]$ \\
        $p_i=1$
        & 485 & 1.0000 & 0.9918
        & $[0.9790,0.9968]$ \\
        \hline
    \end{tabular}
\end{table}

\subsection{Benchmarking the Planning Performance of EGT}
\label{sec:planning_performance}


\noindent We evaluate how EGT-based path selection affects navigation performance on rough terrain. We compare the sampling-based planner in Section~\ref{sec:planner} with GP-navigation~\cite{leininger2024gaussian} and TAL~\cite{tal} on three terrains. For each method, we run 50 trials per terrain. A trial fails if the vehicle gets stuck or tips over. We use the author-provided GP-navigation implementation,
which operates online without pretraining. On the other hand, TAL requires learning a neural transition model, and for this, we use the vehicle transitions extracted from the same simulated execution dataset used to train our FM model.
For navigation, we combine the learned transition model
with adapted $\pi$-MPPI~\cite{pimppi} (400 control sample sequences, horizon 2s). Its cost penalizes pitch
and roll magnitudes and Euclidean distance to the goal.
Note that we use the same $\pi$-MPPI optimization method as the
low-level path-tracking controller in our approach.
Thus, both our approach and TAL use the same underlying
training executions and the $\pi$-MPPI optimization
method, but differ in their prediction targets and
planning objectives. 

\noindent\textbf{Results.}
Table~\ref{tab:navigation_results} reports failure rates using all 50 trials for each method and terrain. Goal-reaching times are computed only over common-success runs, that is, runs in which all three methods succeeded; the common sample counts are 21, 17, and 19 for Terrains~1, 2, and~3, respectively. Our planner has the lowest failure rates on all terrains: 4\% on Terrain~1, 4\% on Terrain~2, and 2\% on Terrain~3. In contrast, GP-navigation fails in 26\%--46\% of trials, and TAL fails in 34\%--48\% of trials.

Over the common-success runs, our planner also has the lowest mean goal-reaching time on each terrain. On Terrain~1, our planner and GP-navigation have similar mean times (27.42\,s and 28.00\,s), but our planner reduces the failure rate from 40\% to 4\%. On Terrains~2 and~3, our planner reduces the mean time by 12.7\% and 20.5\% compared with GP-navigation, while also maintaining lower failure rates. Compared with TAL, our planner reduces the mean goal-reaching time by 36\%--45\% across the three terrains.

\begin{table}[!t]
    \centering
    \caption{\footnotesize{Navigation performance across three terrains,
    with 50 trials per method and terrain. Failure rates use all trials;
    goal-reaching times use common-success runs ($N=21,17,19$ for
    Terrains~1--3). }}
    \label{tab:navigation_results}
    \scriptsize
    \setlength{\tabcolsep}{3pt}
    \renewcommand{\arraystretch}{1.05}
    \begin{tabular*}{\columnwidth}{
        @{\extracolsep{\fill}}llcc@{}
    }
        \toprule
        Terrain & Method
        & \shortstack{Failure rate\\(\%) $\downarrow$}
        & \shortstack{Goal time (s) $\downarrow$\\Mean $\pm$ std.} \\
        \midrule
        Terrain 1
        & Proposed
        & \textbf{4}
        & $\mathbf{27.42} \pm 13.09$ \\
        & GP-navigation
        & 40
        & $28.00 \pm 14.46$ \\
        & TAL
        & 34
        & $49.80 \pm 22.22$ \\
        \midrule
        Terrain 2
        & Proposed
        & \textbf{4}
        & $\mathbf{32.90} \pm 10.65$ \\
        & GP-navigation
        & 46
        & $37.68 \pm 20.69$ \\
        & TAL
        & 48
        & $51.37 \pm 17.19$ \\
        \midrule
        Terrain 3
        & Proposed
        & \textbf{2}
        & $\mathbf{28.76} \pm 7.83$ \\
        & GP-navigation
        & 26
        & $36.18 \pm 13.09$ \\
        & TAL
        & 46
        & $48.82 \pm 15.20$ \\
        \bottomrule
    \end{tabular*}
\end{table}

\section{Conclusions and Future Work}

We presented Execution-Grounded Traversability (EGT) to assess whether a vehicle--controller pair can follow a rough-terrain path. Using closed-loop trajectories predicted by a flow-matching model, EGT computes the probability of satisfying path-tracking and progress constraints. Across 650 simulated trials, EGT achieved 97.4\% classification accuracy, reducing conservative rejection and unsafe acceptance over 3D-pose baselines. The EGT planner achieved 2\%--4\% failure rates and the lowest mean goal-reaching times across three terrains. The real-world validation is presented in Fig.~\ref{fig:teaser_overview} and the accompanying video, which includes experimental comparison with \cite{leininger2024gaussian}. Finally, since our planner accounts for the potential low-level controller's corrective
actions through FM-based execution predictions, it reduces the 
reliance on frequent reference-path re-planning while retaining high-rate feedback control. 

\noindent \textbf{Limitations and Future Work:} Our EGT assessment assumes rough, non-deformable terrains. In the future, we plan to train the FM model on deformable surfaces and condition it on tokens derived from both point clouds and images. Furthermore, embeddings from vision foundation models could be integrated into the FM architecture. In this work, we also assume a fixed vehicle--controller pair. Thus, a promising direction is to enable the FM architecture to take the vehicle embodiment and low-level control structure as additional context. Finally, EGT evaluates execution along the reference path, but it does not check stability past the endpoint. To address this limitation, the planner can replan before the path ends—for example, halfway through—and extend the prediction horizon past the previous endpoint. We are also looking to reduce the computation time of our planner. A major overhead stems from the computation of 3D kinematic prior which in the current work is obtained from solving a non-linear least squares (NLS) problem \cite{manoharan2024bi}. One way to reduce this overhead is to replace the NLS with a learned surrogate.

\bibliographystyle{IEEEtran}
\bibliography{root}
\section{Appendix: FM Network Architecture}\label{appendix}
\subsection{Basic Building Blocks}
\noindent Fig.~\ref{fig:arc_overview} shows the inputs: the reference path
$\boldsymbol{\tau}^{\mathrm{ff}}$, height map
$\mathbf{H}\in\mathbb{R}^{64\times64}$, and validity mask
$\mathbf{M}_{\mathrm{valid}}\in\{0,1\}^{64\times64}$.
The map and mask come from terrain point clouds.
Observed cells have a mask value of one.
Missing cells use zero height.

\noindent\textbf{Multi-Scale Terrain Features:}
An FPN~\cite{lin2017feature} processes both map channels.
A convolutional stem and three downsampling and refinement stages
produce $32\times32$, $16\times16$, and $8\times8$ features
with $64$, $128$, and $256$ channels, respectively.
Top-down fusion and lateral $1\times1$ convolutions give three
grids with $d_{\mathrm{f}}=64$ channels.
We store these as $\mathbf{P}_j\in\mathbb{R}^{d_{\mathrm{f}}\times n_j}$,
where $(n_1,n_2,n_3)=(32^2,16^2,8^2)$.
Each column represents one grid cell.

\noindent\textbf{Path-Wise Terrain Features:}
The $n_y\times n_x$ map covers
$[x_{\min},x_{\max}]\times[y_{\min},y_{\max}]$.
We transform each reference waypoint into the robot-aligned map
frame, normalize it using the map bounds, and clip each coordinate
to $[-1,1]$.
For the resulting location
$\mathbf{q}_k^{\mathrm{map}}=(u_k,v_k)^\top$, the grid coordinates are
$\xi_k=(u_k+1)(n_x-1)/2$ and
$\eta_k=(v_k+1)(n_y-1)/2$.
Bilinear interpolation gives the height $h_k$ in meters,
validity $m_k\in[0,1]$, and features
$\mathbf{f}_{j,k}\in\mathbb{R}^{d_{\mathrm{f}}}$ from each grid.
Each feature grid uses its own resolution.
We combine these samples as
\begin{equation}
    \mathbf{f}_k^{\mathrm{terr}} =\operatorname{col}( \mathbf{f}_{1,k},\mathbf{f}_{2,k}, \mathbf{f}_{3,k},h_k,m_k),
    \label{eq:fm_terrain_features}
\end{equation}
where $\operatorname{col}$ stacks its arguments into a column vector.

\noindent\textbf{Reference-Path Feature:}
For $n_{\mathrm{w}}\geq2$ waypoints, set
$\Delta\mathbf{q}_k^{\mathrm{ff}}
=\mathbf{q}_{k+1}^{\mathrm{ff}}-\mathbf{q}_k^{\mathrm{ff}}$
for $k<n_{\mathrm{w}}-1$, and
$\Delta\mathbf{q}_{n_{\mathrm{w}}-1}^{\mathrm{ff}}=\mathbf{0}$.
The position, displacement, and sequence index form
\begin{equation}
    \mathbf{x}_k^{\mathrm{geom}} =\operatorname{col}\!\left( \mathbf{q}_k^{\mathrm{ff}}, \Delta\mathbf{q}_k^{\mathrm{ff}}, \frac{k}{n_{\mathrm{w}}-1}\right) \in\mathbb{R}^{5}.
    \label{eq:fm_path_geometry}
\end{equation}

\subsection{Constructing the Tokens}
\noindent Each token has $d$ features.
The operator $\operatorname{LN}$ applies layer normalization
to each row.

\noindent\textbf{Spatial Tokens:}
We project the coarsest terrain grid and add learned
grid-position embeddings:
\begin{equation}
    \mathbf{T}_{\mathrm{spatial}} =\operatorname{LN}\!\left( \mathbf{P}_3^\top\mathbf{W}_{\mathrm{spatial}}^\top \right) +\mathbf{E}_{\mathrm{spatial}}^{\mathrm{pos}} \in\mathbb{R}^{n_3\times d},
    \label{eq:fm_spatial_tokens}
\end{equation}
where $\mathbf{W}_{\mathrm{spatial}}
\in\mathbb{R}^{d\times d_{\mathrm{f}}}$ is learned.

\noindent\textbf{Path-Terrain Tokens:}
A two-layer GELU network combines geometry and terrain.
A validity gate controls the terrain contribution:
\begin{align}
    \mathbf{u}_k^{\mathrm{terr}} &=\operatorname{MLP}_{\mathrm{fuse}}\!\left( \operatorname{col}( \mathbf{x}_k^{\mathrm{geom}}, \mathbf{f}_k^{\mathrm{terr}}) \right),
    \label{eq:fm_terrain_geometry_fusion}\\
    \mathbf{g}_k &=(0.1+0.9m_k)\, \sigma(\mathbf{w}_{\mathrm{gate}}m_k +\mathbf{b}_{\mathrm{gate}}).
    \label{eq:fm_validity_gate}
\end{align}
The gate parameters are learned, and $\sigma$ is the
elementwise sigmoid.
At $m_k=0$, each gate entry is at most $0.1$.
The geometry-only branch remains active:
\begin{equation}
    \begin{aligned}
        \widetilde{\mathbf{t}}_{\mathrm{way},k} &=\mathbf{W}_{\mathrm{geom}}\mathbf{x}_k^{\mathrm{geom}} +\mathbf{g}_k\odot\mathbf{u}_k^{\mathrm{terr}} +\mathbf{e}_k^{\mathrm{pos}},\\
        \mathbf{t}_{\mathrm{way},k} &=\operatorname{Dropout}\!\left( \operatorname{LN}( \widetilde{\mathbf{t}}_{\mathrm{way},k}) \right).
    \end{aligned}
    \label{eq:fm_waypoint_tokens}
\end{equation}
Here, $\mathbf{W}_{\mathrm{geom}}\in\mathbb{R}^{d\times5}$
and $\mathbf{e}_k^{\mathrm{pos}}$ are learned.
The operator $\odot$ denotes elementwise multiplication.
We stack the tokens in path order as rows of
$\mathbf{T}_{\mathrm{way}}\in\mathbb{R}^{n_{\mathrm{w}}\times d}$.
We disable dropout at inference.

\noindent\textbf{Kinematic, Reference, and Goal Tokens:}
We compute pitch, roll, height, and twelve wheel--ground
contact point coordinates (for 4 wheels) along the path using~\cite{manoharan2024bi}.
A three-layer transformer encoder with sinusoidal positional
embeddings gives
$\mathbf{T}_{\mathrm{kin}}\in\mathbb{R}^{(h+1)\times d}$.
The reference latent and the goal in the robot's local frame give 
$\mathbf{t}_{\mathrm{ref}} =\mathbf{W}_{\mathrm{ref}}\operatorname{LN}(\mathbf{z}^{\mathrm{ff}}), \quad \mathbf{t}_{\mathrm{goal}} =\mathbf{W}_{\mathrm{goal}}\operatorname{LN}(\mathbf{g}_{\mathrm{goal}}).$

\subsection{Latent-Flow Transformer}
\label{app:flow_transformer}
\noindent We split $\mathbf{z}_r\in\mathbb{R}^{32}$ into 16 pairs.
We concatenate each pair with the same 64-dimensional
sinusoidal embedding of flow time $r$.
A shared $66\rightarrow256$ projection, layer normalization,
and group-position embeddings form
$\mathbf{T}_{\mathrm{lat}}\in\mathbb{R}^{16\times256}$.

Let $\overline{\mathbf{T}}_{\mathrm{lat}}
=\operatorname{LN}(\mathbf{T}_{\mathrm{lat}})$ and
$\overline{\mathbf{C}}=\operatorname{LN}(\mathbf{C})$.
In multi-head attention
$\operatorname{MHA}$~\cite{vaswani2017attention},
the first argument supplies queries.
The second supplies keys and values.
The network computes
\begin{align}
    \mathbf{T}^{(0)} &=\mathbf{T}_{\mathrm{lat}} +0.5\,\operatorname{MHA}( \overline{\mathbf{T}}_{\mathrm{lat}}, \overline{\mathbf{C}}),
    \label{eq:latent_conditioning}\\
    \mathbf{T}^{(b)} &=\operatorname{Enc}_b(\mathbf{T}^{(b-1)}), \quad b=1,\ldots,4,
    \label{eq:latent_encoder}\\
    \mathbf{v}_{\theta}(\mathbf{z}_r,r,\mathbf{C}) &=\operatorname{flatten}\!\left( \operatorname{Head}(\mathbf{T}^{(4)})\right).
    \label{eq:latent_output}
\end{align}
Each pre-normalized encoder uses
self-attention and a $256\rightarrow512\rightarrow256$
GELU network with residual connections.
Residual-branch dropout is $0.1$ during training.
All attention modules use four heads.
Only the initial cross-attention uses $\mathbf{C}$.
The shared $256\rightarrow2$ head produces two values per token.
Flattening the rows in latent-group order gives the
32-dimensional output.

\section{AI Acknowledgment}
GPT6-Astra and Gemini4-Argon were used for language editing of this paper.

\end{document}